\documentclass[journal]{IEEEtran}
\usepackage{ifpdf}

\usepackage{cite}

\ifCLASSINFOpdf
   \usepackage[pdftex]{graphicx}
   \usepackage[pagebackref=true,breaklinks=true,colorlinks,bookmarks=false]{hyperref}
\else
\fi
\usepackage{amsmath}
\usepackage[pagebackref=true,breaklinks=true,colorlinks,bookmarks=false]{hyperref}

\usepackage{algorithmic}
\usepackage[pagebackref=true,breaklinks=true,colorlinks,bookmarks=false]{hyperref}

\usepackage{array}
\usepackage[pagebackref=true,breaklinks=true,colorlinks,bookmarks=false]{hyperref}

\usepackage{url}
\usepackage[pagebackref=true,breaklinks=true,colorlinks,bookmarks=false]{hyperref}

\begin{document}
%
\title{DeepIlluminance: Contextual Illuminance Estimation via Deep Neural Networks}
%
%
%

\author{Jun~Zhang,
        Tong~Zheng,
        Shengping~Zhang,
        and~Meng~Wang
\thanks{J. Zhang, T. Zheng, and M. Wang are with the School of Computer Science and Information Engineering, 
Hefei University of Technology, Hefei, Anhui, 230601 China.}
\thanks{S. Zhang is with the School of Computer Science and Technology, Harbin Institute of Technology, Weihai, Shandong, 264209 China.}
\thanks{Corresponding author: Jun Zhang (e-mail: zhangjun1126@gmail.com)}}

\maketitle

\begin{abstract}
Computational color constancy refers to the estimation of the scene illumination and makes the perceived color relatively stable under varying illumination.
In the past few years, deep Convolutional Neural Networks (CNNs) have delivered superior performance in illuminant estimation.
Several representative methods formulate it as a multi-label prediction problem by learning the local appearance of image patches using CNNs.
However, these approaches inevitably make incorrect estimations for the ambiguous patches affected by their neighborhood contexts.
Inaccurate local estimates are likely to bring in degraded performance when combining into a global prediction.
To address the above issues, we propose a contextual deep network for patch-based illuminant estimation equipped with refinement.
First, the contextual net with a center-surround architecture extracts local contextual features from image patches, and generates initial illuminant estimates and the corresponding color corrected patches.
The patches are sampled based on the observation that pixels with large color differences describe the illumination well.
Then, the refinement net integrates the input patches with the corrected patches in conjunction with the use of intermediate features to improve the performance.
To train such a network with numerous parameters, we propose a stage-wise training strategy, 
in which the features and the predicted illuminant from previous stages are provided to the next learning stage with more finer estimates recovered.
Experiments show that our approach obtains competitive performance on two illuminant estimation benchmarks.
\end{abstract}

\begin{IEEEkeywords}
Illuminant estimation, color constancy, local context, refinement, deep convolutional neural networks.
\end{IEEEkeywords}

%

\section{Introduction}
The computational color constancy~\cite{Gijsenij11} aims to estimate the unknown color of the illuminating light source given an RGB image, and then correct the chromaticity of the light source using the illuminant estimate.
It is inherently ambiguous and a technically ill-posed problem because both the spectral distribution of the illuminant and the scene reflectance are unknown.
But it has been attracting increasing interest in the vision communities since various high-level visual understanding tasks require discounting the illuminant to obtain the ``true color'' or reflectance of objects, such as material recognition~\cite{Wang18}. 

Early approaches are derived from image statistics or physical models that make a variety of assumptions about the image, such as gray-world~\cite{Barnard02}, white-patch~\cite{Gijsenij07}, and Lambertian surface~\cite{Finlayson01}.
These methods also assume that the illuminant in the image is spatially uniform. 
These assumptions are coarse approximations to the real-world cases and limit the performance.

To improve these methods, another line of research learns a discriminative objective function based on hand-crafted features to estimate the scene illumination directly by machine learning techniques, such as neural networks~\cite{Cardei02}, support vector regression~\cite{Xiong06}, Bayesian estimation~\cite{Gehler08}, and exemplar learning~\cite{Joze14}.
However, these methods fail to estimate the scenes where the colors of objects are inherently similar to those of the light sources.  

Very recently, Convolutional Neural Networks (CNNs) have been employed to learn the relationship between the pixels and the chromaticity of the light source, which outperform previous methods by a wide margin~\cite{Bianco17,Hu17,Barron17,Shi16,Bianco15,Barron15,Lou15} and can be roughly grouped as global approaches and local approaches.
The state-of-the-art performances~\cite{Hu17,Barron17} are obtained by considering the semantic information at a global level and exploiting different model components on different datasets.
In addition, what is often overlooked among local approaches is that color constancy is achieved by taking contextual information between the image patch and its surrounding illumination into account~\cite{Hansen07}.
Since small patches can be greatly affected by surrounding variance, calculating local estimates can be difficult and the total global prediction suffers from the degraded performance. 

\begin{figure*}[!t]
\centering
   \includegraphics[width=1\linewidth]{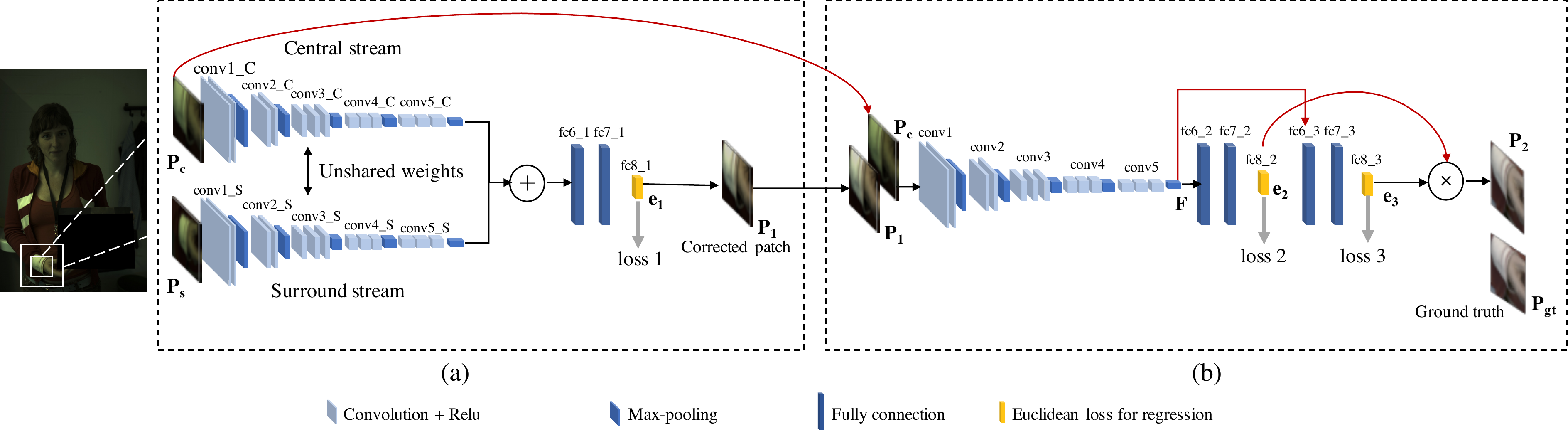}
   \caption{The pipeline of our approach. (a) The corrected patch is generated from the contextual net with a center-surround architecture. The refinement net is connected to refine the initial estimate with skip connections from the input patch and intermediate features.}
\label{fig:framework}
\end{figure*}

To address the aforementioned issues, we propose a novel patch-based deep network with local contextual information for illuminant estimation and refinement.
First, rather than introducing random or uniform sampling strategies~\cite{Bianco15,Bianco17,Hu17}, we sample image patches based on the selected bright and dark pixels from the color image by calculating and ranking the projection distances of all color pixels to the mean vector in the RGB domain.
The rational behind our method is that pixels with large color differences can well describe the illumination direction~\cite{Cheng14}.
The sampled patches are taken as inputs to our network, which is comprised of two cooperative sub-networks based on the VGG-16 architecture~\cite{Simonyan14}: a feedforward contextual net and a refinement net, as shown in Figure~\ref{fig:framework}.
The contextual net exploits both local features and neighbor contextual features to generate an initial illuminant estimate and correct the input patch via the diagonal transform~\cite{vonKries70}.
The refinement net efficiently learns features over the joint input-output space by stacking the corrected patch and the original patch, and uses the intermediate features encoded in itself with skip connections to generate a finer illuminant estimation.
Our approach allows for reevaluation of the illuminant color and features across the sampled patches.
It is similar in spirit to some structured prediction methods~\cite{Newel16,Li16}, which have made successive predictions with intermediate supervision to refine predictions.

Inspired by the conclusion~\cite{Barshan15} that the stage-wise training can avoid gradient diffusion and overfitting for deep networks by decoupling the feature extraction layers from the classification layer across successive stages, we propose a stage-wise training strategy by breaking down the entire network into two related sub-tasks, in which the predicted illuminant and the intermediate features are passed through the networks stage-by-stage.
Finally, to obtain a global illuminant estimate, we use median pooling on all the local estimates.
We show experimentally that the proposed framework achieves competitive performance over several state-of-the-art methods on two illuminant estimation benchmarks.

The main contributions of this work are listed as follows.
\begin{itemize}
  \item We propose a novel contextual deep network using a center-surround architecture with a refinement mechanism for illuminant estimation, which captures local contextual features for initial estimation and reevaluates the features in the joint input-output space used in conjunction with intermediate supervision for finer estimation.
 The proposed approach can be viewed as the first piece of work that shows how contextual information and successive refinement are critical to improve  illuminant estimation, even without the use of semantic information~\cite{Barron17,Hu17}.
  \item We sample image patches by selecting bright and dark pixels with large color differences in the RGB space.
    To the best of our knowledge, this is the first work to sample patches directly from the color domain for illuminant estimation.
  \item We propose a stage-wise training strategy to leverage the initial estimation and intermediate supervision from the illuminant color, which serves to increase efficiency and reduce memory usage of our network while improving precision of illuminant estimation.
\end{itemize}

We review related work in Section~\ref{work} and present our approach in Section~\ref{approach}, focusing on our contextual network architecture, refinement, and our training procedure. 
We present experiments and results in Section~\ref{experiments}.
In Section~\ref{conclusion} discussion and further perspectives of this work are presented.

\section{Related work}
\label{work}

Prior to the deep learning revolution, color constancy algorithms mainly relied on different assumptions and handcrafted features~\cite{Weijer07,Finlayson11}.
In this section, we restrict ourselves mostly to recent methods that exploit CNNs.
These methods can be roughly grouped as local approaches and global approaches. 

\subsection{Local approaches}
An early attempt~\cite{Bianco15} using CNNs is to extract {\it conv} features of non-overlapping patches based on AlexNet~\cite{Krizhevsky12}, and pass them to a support vector regression to estimate the illuminant color.
To deal with non-uniform illumination, a multiple illuminant detector~\cite{Bianco17} using a kernel density estimator is proposed to determine if the image contains single or multiple illuminants.
To handle the ambiguities of unknown reflections and local patch appearances, a deep specialized network~\cite{Shi16} is presented where a hypotheses network generates two hypotheses of illuminants for a UV patch and a selection network adaptively picks the confident estimations from these hypotheses.
Another noteworthy work~\cite{Oh17} clusters the illuminants and then feeds them into a CNN with the new illumination labels.
The final illuminant color is estimated by computing the weighted average of the cluster centers.
These methods conduct the prediction in a large and diverse hypothesis space given limited training samples, making the results still unsatisfactory.
More recently, an end-to-end fully convolutional network~\cite{Hu17} is proposed to produce local estimates followed by a confidence weight pooling to generate the global prediction.
This method implicitly takes advantage of human faces as high-confidence regions and achieves impressive performance based on two backbone models.
In this work, we propose a CNN-based framework to take the contextual information into consideration and refine local estimates of image patches in the joint input-output space.

\subsection{Global approaches}
Global illumination estimation based on the whole images has been addressed by ~\cite{Lou15,Barron15,Barron17}.
In the work of ~\cite{Lou15}, three stacked CNNs are trained sequentially to get hierarchy features of the full image for illuminant estimation. 
Barron~\cite{Barron15} and Barron \& Tsai~\cite{Barron17} formulated the task as a 2D spatial localization problem by learning {\it conv} filters in the log-chroma plane.
These methods consider the semantic information at a global level, and predict the illuminant color with local details lost.
In contrast, our work estimates the illuminant colors at a patch level by sampling semantically valuable local regions.

\section{Network architecture}
\label{approach}

We begin by describing the contextual net for initial estimation in Section~\ref{master}, followed by a detailed description of our refinement net in Section~\ref{refine}.
Finally, we present the stage-wise training method in Section~\ref{train}. 

\subsection{Contextual network for initial estimation}
\label{master}
We choose the VGG-16 network~\cite{Simonyan14} as our backbone model, which is pre-trained on the ImageNet dataset~\cite{Russakovsky15} for object recognition.
We replace the last layer of $1000$ units that predicts the ImageNet classes with a layer containing $3$ units, encoding the continuous illuminant color for RGB channels.
In principle, the backbone model can be replaced by other advanced shallow or deep networks in our system.

As illustrated in Figure~\ref{fig:framework}(a), the contextual net consists of two central and surround separate streams, one fusion layer, and one decision net.
The central stream takes an image patch $\mathbf{P_c}$ comprising selected bright and dark pixels as input and extracts its local features.
The surround stream takes the surrounding neighbor $\mathbf{P_s}$ of the central patch and extracts global features to provide the larger contextual information.
The kernel weights of the two streams are unshared.
The fusion layer combines the last {\it conv} features from the two streams via element-wise summation, and gives it to the top decision net that consists of three fully connected layers separated by a ReLU layer for the initial local estimation $\mathbf{e_1}$.
Finally, the initial estimate is used to generate the color corrected patch $\mathbf{P_{1}}$ by applying the diagonal transform~\cite{vonKries70} to the original patch $\mathbf{P_c}$.
The working of the contextual net can be mathematically described by the following equations:
\begin{equation}
\begin{aligned}
\mathbf{e_1} & =\mathcal{R}\left (\mathcal{F}(\mathbf{P_c};\mathbf{W_c}) + \mathcal{F}(\mathbf{P_s};\mathbf{W_s}); \mathbf{W_1}\right ) \\
\mathbf{P_1} &=\mathcal{M}\left (\mathbf{e_1}, \mathbf{P_c} \right )\\
\end{aligned}
\end{equation}
where $\mathcal{F(\cdot)}$ denotes the output feature maps generated by the {\it conv} blocks of the two streams with weights $\mathbf{W_c}$ and $\mathbf{W_s}$, $\mathcal{R(\cdot)}$ denotes illuminant regression by the {\it fc} layers with parameter $\mathbf{W_1}$, and $\mathcal{M(\cdot)}$ denotes the diagonal transform.

One reason for the usage of such a center-surround architecture is that multi-scale information is known to be important in capturing spatial context in color constancy~\cite{Hansen07,Hurlbert07}.
Furthermore, by considering the central patch twice (i.e. in both the central stream and the surround stream) we implicitly put more focus on the pixels closer to the center of a patch, which can also improve precision of illuminant estimation.
Note that the proposed contextual network shares the similar structure with the pseudo-siamese network proposed in~\cite{Zagoruyko15} except that stream outputs are summed.
As verified in experiments (Section~\ref{ablation}), the proposed architecture improves the illuminant estimation accuracy compared with other variants.

\subsection{Refinement network for finer prediction}
\label{refine}
We also adopt the VGG-16 network as our fundamental building block for illuminant refinement.
We take the network architecture further by stacking the refinement net and feeding the output of the contextual net as input, 
which provides the network with a mechanism for successive bottom-up processing in the joint input-output space and allows for the use of the intermediate features encoded in the refinement net.

Figure~\ref{fig:framework}(b) shows a detailed illustration of the refinement net, which concatenates the corrected patch $\mathbf{P_{1}}$ and the original patch $\mathbf{P_c}$ along the third dimension with the long skip connection to generate new intermediate feature maps $\mathbf{F}=\mathcal{F}\left (\mathcal{CAT}\left ( \mathbf{P_c},\mathbf{P_1} \right );\mathbf{W_2} \right )$ and the corresponding illuminant estimate $\mathbf{e_2}$, where $CAT$ indicates the concatenation operator.
Then, we append extra three fully-connected layers (denoted as {\it $fc6\_3$}, {\it $fc7\_3$}, and {\it $fc8\_3$}) behind the fifth {\it conv} block, which allows the intermediate features to be processed again to further predict the illuminant color $\mathbf{e_3}$ on the original patch.
Finally, the new estimate is combined with $\mathbf{e_2}$ via element-wise product to improve the scale of the illuminant values, and even correct the wrong estimate to generate the final corrected patch $\mathbf{P_2}$:
\begin{equation}
\begin{aligned}
\mathbf{P_2} =\mathcal{M}\left (\mathbf{e_2}\circ \mathcal{R}\left (\mathbf{F}; \mathbf{W_3} \right ), \mathbf{P_c} \right )
\end{aligned}
\end{equation}
where $\circ$ is an element-wise product.

Our method bears some similarity to structured prediction methods~\cite{Newel16,Li16} in a broad sense of being successive predictions included with the input, but our work is tailored for illuminant estimation and the network is different.
Such a refinement implicitly ensures consistency of the output with the input by serving the corrected patch as an illuminant prior, and provides richer intermediate supervision and helps learn stage-specific refinements to predict the illuminant color.
It is important to note that the {\it fc} weights are not shared across networks, and three losses are applied to the predictions separately using the same ground truth. 
The details for the training procedure are described below.

\subsection{Training and implementation}
\label{train}
The entire network is trained using ground-truth RGB labels, i.e. to perform illuminant regression.
We use a Euclidean loss with the following learning objective function:
\begin{equation}
\frac{1}{N}\sum_{i=1}^{N} min\left ( \left \| \mathbf{e}_{i}-\mathbf{e}_{i}^{\ast} \right \|_{2}^{2} \right )
\end{equation}
where $N$ is the number of training samples in a batch, $\mathbf{e}_{i}$ is the illuminant estimate for the $i$-th patch, and $\mathbf{e}_{i}^{\ast}$ is the corresponding ground-truth illuminant color.

\begin{figure*}[!t]
\centering
   \includegraphics[width=1\linewidth]{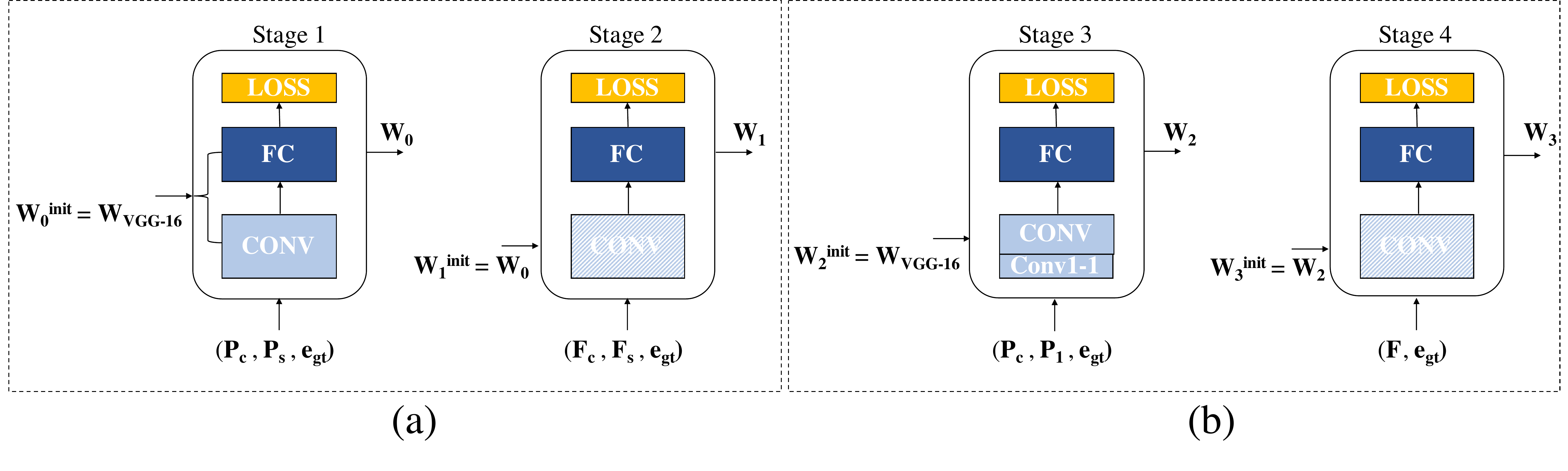}
   \caption{The schematic diagram of the stage-wise training with information evolution. Note that only the {\it conv} layers are initialized using the previous stage. (a) The context net. (b) The refinement net. The boxes with patter fills represent the fixed weights with a learning rate of 0.}
\label{fig:training}
\end{figure*}

Our approach reduces illuminant estimation to a sequence of predictions.
We propose a stage-wise training strategy shown in Figure~\ref{fig:training}, in which each stage is trained separately so that the features and the predicted illuminant at the early stages are provided to the next learning stage for finer estimation.
Stage-wise training has been previously studied~\cite{Barshan15}, and shown substantially improved performance in pose estimation~\cite{Ramakrishna14}.
\begin{itemize}
\item Stage 1: the weights of the {\it $fc8\_1$} layer are initialized with zero-mean Gaussians.
Other layers are initialized with the pre-trained VGG-16 network~\cite{Simonyan14}.
The central and the surround streams are fine-tuned to estimate the illuminant, respectively.
\item Stage 2: we only train the weights in the {\it fc} layers on top of the combination of output features $\mathbf{F_c}$ and $\mathbf{F_s}$ to generate an initial illuminant estimate $\mathbf{e_1}$ and its corrected patch $\mathbf{P_1}$.
The weights in the {\it conv} layers are transferred from the preceding stage and fixed at this stage.
\item Stage 3: the {\it conv1-1} layer and the {\it fc} layers of the refinement net are initialized with zero-mean Gaussians.
Other convolutional parameters are assigned with the VGG-16 weights.
We train this stage with the contextual net fixed and predict the illuminant $\mathbf{e_2}$ by stacking the original patch $\mathbf{P_c}$ and its corrected patch $\mathbf{P_1}$.
\item Stage 4: at the second refinement stage, we train the output layers (i.e. {\it $fc6\_3$}, {\it $fc7\_3$} and {\it $fc8\_3$}) based on the intermediate features $\mathbf{F}$ obtained from the stage 3 to produce a new estimate $\mathbf{e_3}$, keeping the weights of all other layers learned from the preceding stages fixed.
\end{itemize}

We also try to train the whole network in an end-to-end manner. 
However, it is hard to train such a deep network with insufficient data in the benchmark datasets. 
As reported in the experiments, we are able to achieve good solutions in a reasonable computation time using the stage-wise training strategy.

{\bf Implementation details:}
We implement the network based on the Caffe framework~\cite{Jia14} on a single NVIDIA Titan XP GPU. 
The standard stochastic gradient descent (SGD) is employed for optimization, where the initial learning rate, the momentum and the weight decay are set to $0.001$, $0.9$ and $0.0005$, respectively.
We set the batchsize to $23$ and the maximum iteration step to $160K$, and decay the learning rates by a factor of $0.1$ every $50K$ iterations. 
The proposed network takes only $0.16$s to estimate the illuminant color of a whole image.
The source code is released at \url{https://github.com/pencilzhang/DeepIlluminance-computational-color-constancy.git}.

\section{Experimental results}
\label{experiments}
In this section, we experimentally evaluate the proposed approach, and compare with state-of-the-art methods on single illuminant estimation.

\subsection{Setup}
\subsubsection{Datasets and preprocessing}
We use the reprocessed Color Checker Dataset~\cite{Gehler08} and the NUS 8-camera dataset~\cite{Cheng14} for benchmarking.
The reprocessed Color Checker Dataset contains $568$ raw images.
The NUS 8-camera dataset consists of $8$ subsets captured by $8$ different cameras, where each subset contains $210$ images.
For both datasets, the Macbeth Color Checker chart is placed in each image to estimate the ground-truth illuminant color and masked out for both training and testing. 
Following previous work~\cite{Gijsenij11}, three-fold cross validation is used to evaluate the network on both datasets.
Since the VGG-16 network is pre-trained on the ImageNet dataset, where images are gamma-corrected for display, we also apply a gamma correction of $\gamma = 1/2.2$ on linear RGB images.

\subsubsection{Patch sampling}
We project all the pixels of an image onto the mean vector and then rank the projection distances according to the method in~\cite{Cheng14}.
The pixels ranking in the top $d\%$ distance are selected as bright pixels while the bottom $d\%$ are dark pixels.
Then, we randomly sample $M$ central patches of $224 \times 224$ size containing both the bright and dark pixels and their surrounding patches with $2$ times the size of the central patches as inputs to the network.
In fact, we found that we could not obtain better accuracy and higher coverage with smaller or larger patch size.
In this work, we set $M=15$ and $d=\left \{ 3.5,5,10 \right \}$ in ascending order until the number of the sampled patches meets the quantity requirement.
By taking patches as inputs, we have a much larger number of training samples to train the network.

\subsubsection{Metrics}
We adopt the angular error $\epsilon$ between the estimated illumination $e$ and the ground truth $e^{\ast}$ as the performance measure:
\begin{equation}
\epsilon=arccos\left ( \frac{e\cdot e^{\ast}}{\left \| e \right \|\cdot \left \| e^{\ast} \right \|} \right )\end{equation}
We report the mean, median, tri-mean, means of the lowest-error $25\%$ and highest-error $25\%$ of the angular error as evaluation metrics.
In addition, we use the $95$ percentile for the reprocessed Color Checker dataset. 
We run $8$ different experiments on the subset of the NUS 8-camera dataset and evaluate with the geometric mean (Geomean).

\subsection{Ablation studies}
\label{ablation}
We analyze the contribution of the model components on the reprocessed Color Checker Dataset and summarize the results in the following section. 

\subsubsection{Influence of patch sampling}
We compare the illuminant estimation results of our patch sampling method and the random sampling method. 
Both methods take the same size and number of patches as inputs.
Since the refinement net has a substantial effect on the performance of local estimation, we only report results of the contextual net in this experiment.
Table~\ref{patch} summarizes the quantitative results, which show our method achieves considerable improvements over the random sampling method (`RS') in terms of all metrics.
\begin{table}[t!]
\renewcommand{\arraystretch}{1.3}
\caption{Performance comparison of our patch sampling method by selecting the bright and dark pixels against random sampling}
\label{patch}
\centering
\begin{tabular}{lrrrrrr}
\hline
Method & Mean & Med & Tri & \multicolumn{1}{c}{\begin{tabular}[c]{@{}c@{}}Best\\ 25\%\end{tabular}} & \multicolumn{1}{c}{\begin{tabular}[c]{@{}c@{}}Worst\\ 25\%\end{tabular}} & \multicolumn{1}{c}{\begin{tabular}[c]{@{}c@{}}95th \\ Pct \end{tabular}} \\
\hline
Ours & 1.94 & 1.29 & 1.47 & 0.37 &  4.56 & 5.74\\
RS & 2.24 & 1.57 & 1.68 & 0.52 & 5.25 & 6.82 \\
\hline
\end{tabular}
\end{table}

\begin{figure*}
\begin{center}
   \includegraphics[width=1\linewidth]{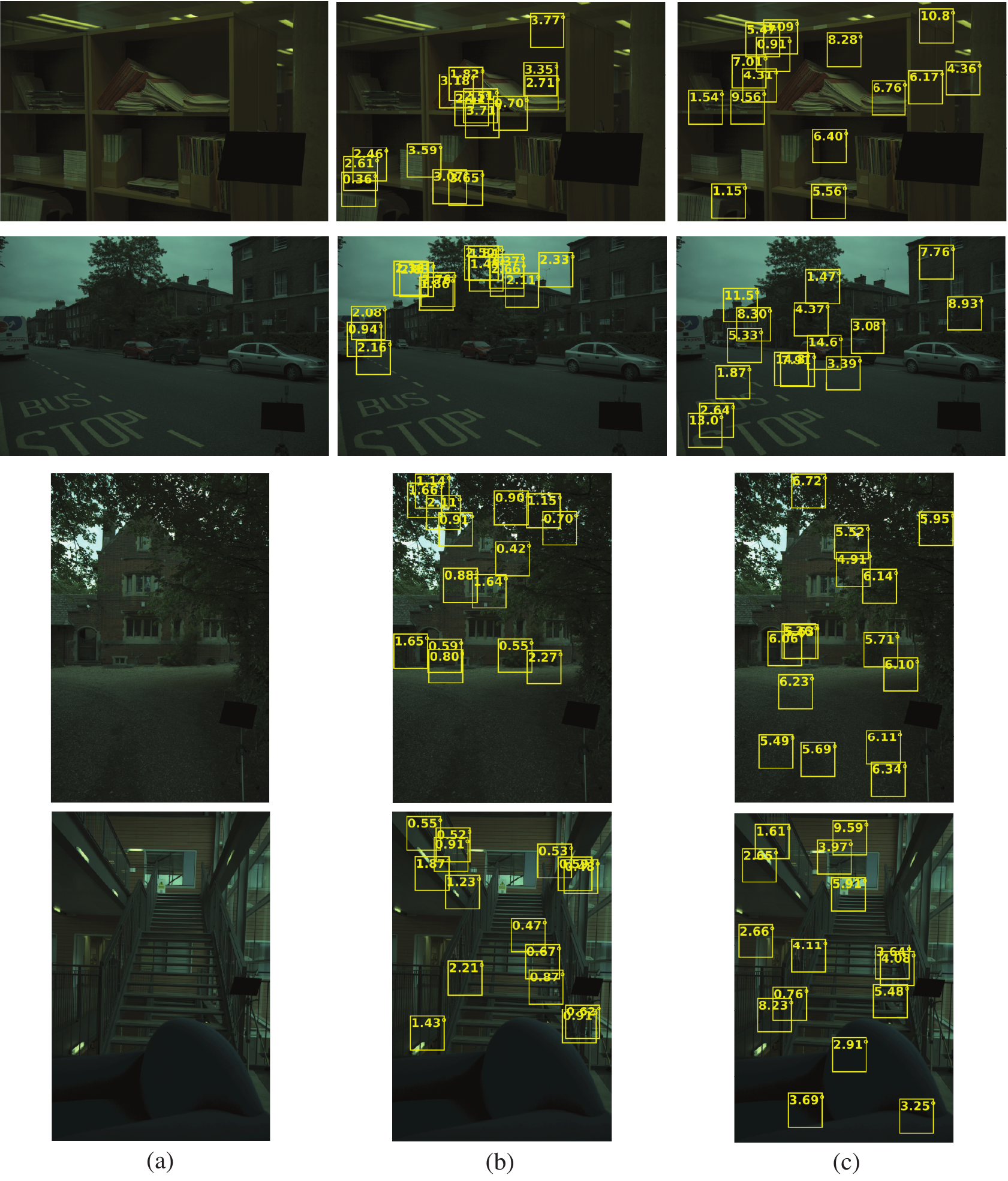}
\end{center}
   \caption{Examples with sampled patches using different sampling methods. (a) Input image. (b) Our method. (c) Random sampling. The angular error is provided at the top of each patch.}
\label{fig:patch}
\end{figure*}

Figure~\ref{fig:patch} shows two examples with sampled patches obtained by the two methods and their corresponding angular errors.
We can observe that the sampled patches based on our method can be estimated more accurately, and are more coherent with the actual brightness contrast pattern in the image, showing large color differences between pixels.

\subsubsection{Effect of the center-surround architecture}
To verify the necessity of our center-surround architecture in the contextual network, we compare five design choices in this experiment,  including single central network, 2-channel (stacking the central and surrounding patches as a 2-channel image), Siamese, Pseudo-siamese (the same architecture with Siamese but with unshared weights), and our proposed contextual architecture.
Results are reported in Table~\ref{context}. 
\begin{table}[t!]
\renewcommand{\arraystretch}{1.3}
\caption{Performance comparison of different model variations on the reprocessed Color Checker dataset}
\label{context}
\centering
\begin{tabular}{lrrrrrr}
\hline
Method & Mean & Med & Tri & \multicolumn{1}{c}{\begin{tabular}[c]{@{}c@{}}Best\\ 25\%\end{tabular}} & \multicolumn{1}{c}{\begin{tabular}[c]{@{}c@{}}Worst\\ 25\%\end{tabular}} & \multicolumn{1}{c}{\begin{tabular}[c]{@{}c@{}}95th \\ Pct\end{tabular}} \\
\hline
Central & 2.16 & 1.52 & 1.66 & 0.44 &  5.17 & 6.43 \\
Context (2-channel) & 2.21 & 1.63 & 1.81 & 0.51 & 5.21 & 7.02 \\
Context (siamese) & 2.37 & 1.68 & 1.84 & 0.65 & 5.35 & 6.99 \\
Context (pseudo-siamese) & 1.99 & 1.41 & 1.52 & 0.40 & 4.59 & 5.82 \\
Context (ours) & \textbf{1.94} & \textbf{1.29} & \textbf{1.47} & \textbf{0.37} & \textbf{4.56} & \textbf{5.74} \\
\hline
\end{tabular}
\end{table}
From the results, we can see that the architectures with unshared weights (i.e. pseudo-siamese and ours) gives a significant boost to other models, and our model exhibits clearly the best performance among all models.

\begin{figure*}
\begin{center}
   \includegraphics[width=1\linewidth]{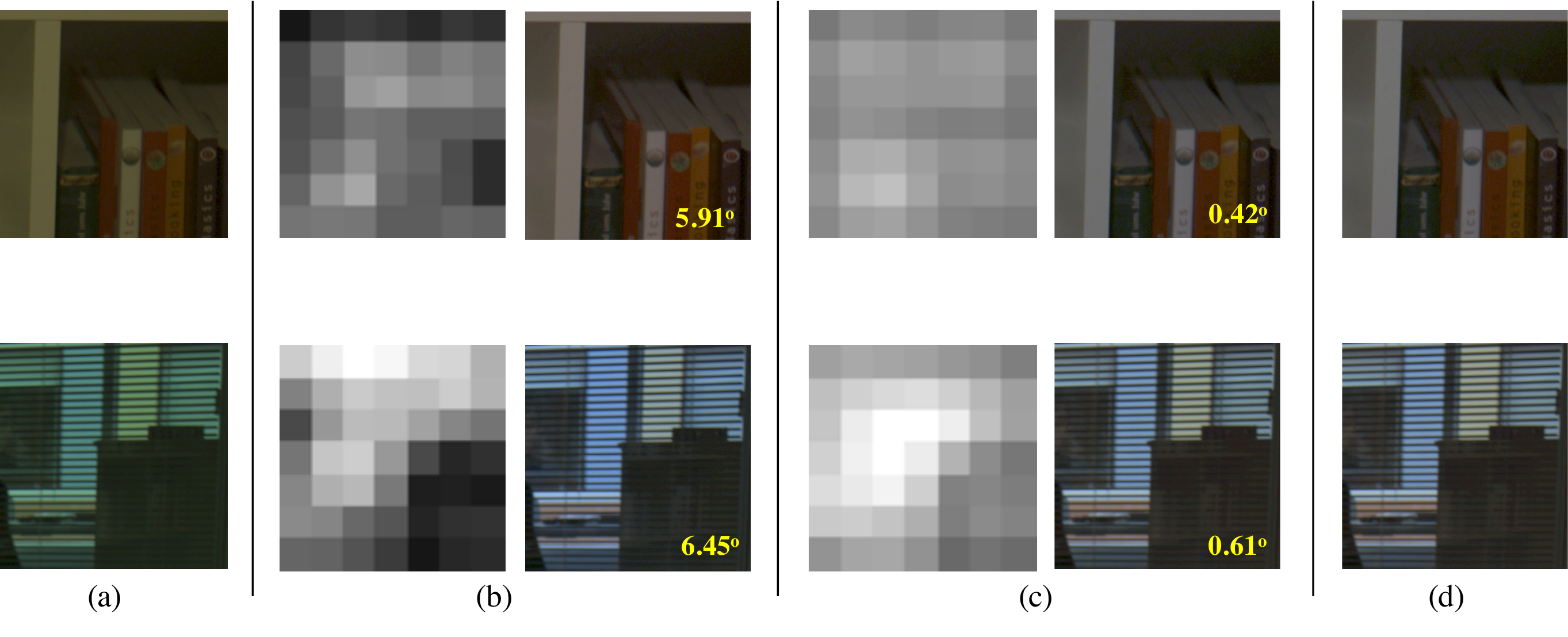}
\end{center}
   \caption{Comparison of representative feature maps of local patches obtained from the single central stream and the proposed contextual network. (a) Input image. (b) Pooled {\it conv5} feature maps and the restored patches obtained from the central stream. (c) Fused feature maps and the restored patches obtained from the proposed network. (d) Ground truth. The angular error is provided at the bottom of each patch.}
\label{fig:vis}
\end{figure*}

To further understand what information is better preserved by the contextual representation, we compare the fused feature maps with the pooled {\it conv5} feature maps of the single central stream, as shown in Figure~\ref{fig:vis}.
We observe that the central stream alone is influenced by larger pixel intensity in the patches while the proposed contextual network can co-adjust the features of the central and surrounding patches.

\subsubsection{Effect of the refinement}
To better understand how our refinement components benefit the performance, we perform a detailed comparison of their performance.
Table~\ref{component} shows the performance of several versions of refinement.
\begin{table}[t!]
\renewcommand{\arraystretch}{1.3}
\caption{Performance comparison of different model variations on the reprocessed Color Checker dataset} 
\label{component}
\centering
\begin{tabular}{lrrrrrr}
\hline
Method & Mean & Med & Tri & \multicolumn{1}{c}{\begin{tabular}[c]{@{}c@{}}Best\\ 25\%\end{tabular}} & \multicolumn{1}{c}{\begin{tabular}[c]{@{}c@{}}Worst\\ 25\%\end{tabular}} & \multicolumn{1}{c}{\begin{tabular}[c]{@{}c@{}}95th \\ Pct\end{tabular}} \\
\hline
Context & 1.94 & 1.29 & 1.47 & 0.37 & 4.56 & 5.74 \\
Refine ($\mathbf{e_2}$) & 1.85 & 1.14   & 1.37    & 0.45      & 4.52       & 5.60            \\
Refine ($\mathbf{e_2}+\mathbf{e_3}$) & \textbf{1.82} & \textbf{1.11}   & \textbf{1.29}    & \textbf{0.35}      & \textbf{4.36}       & \textbf{5.44}            \\
\hline
\end{tabular}
\end{table}
With the corrected patch stacked with the original patch, the performance of the subsequent refinement with $\mathbf{e_2}$ increases over the contextual network, which indicates that encompassing both the input and the initial estimate can expand the expressive power of hierarchical features over the joint space.
We further see that the intermediate supervision in the further refinement with $\mathbf{e_3}$ does offer an improvement to the final illuminant estimation performance.

\begin{figure*}[t!]
\begin{center}
   \includegraphics[width=1\linewidth]{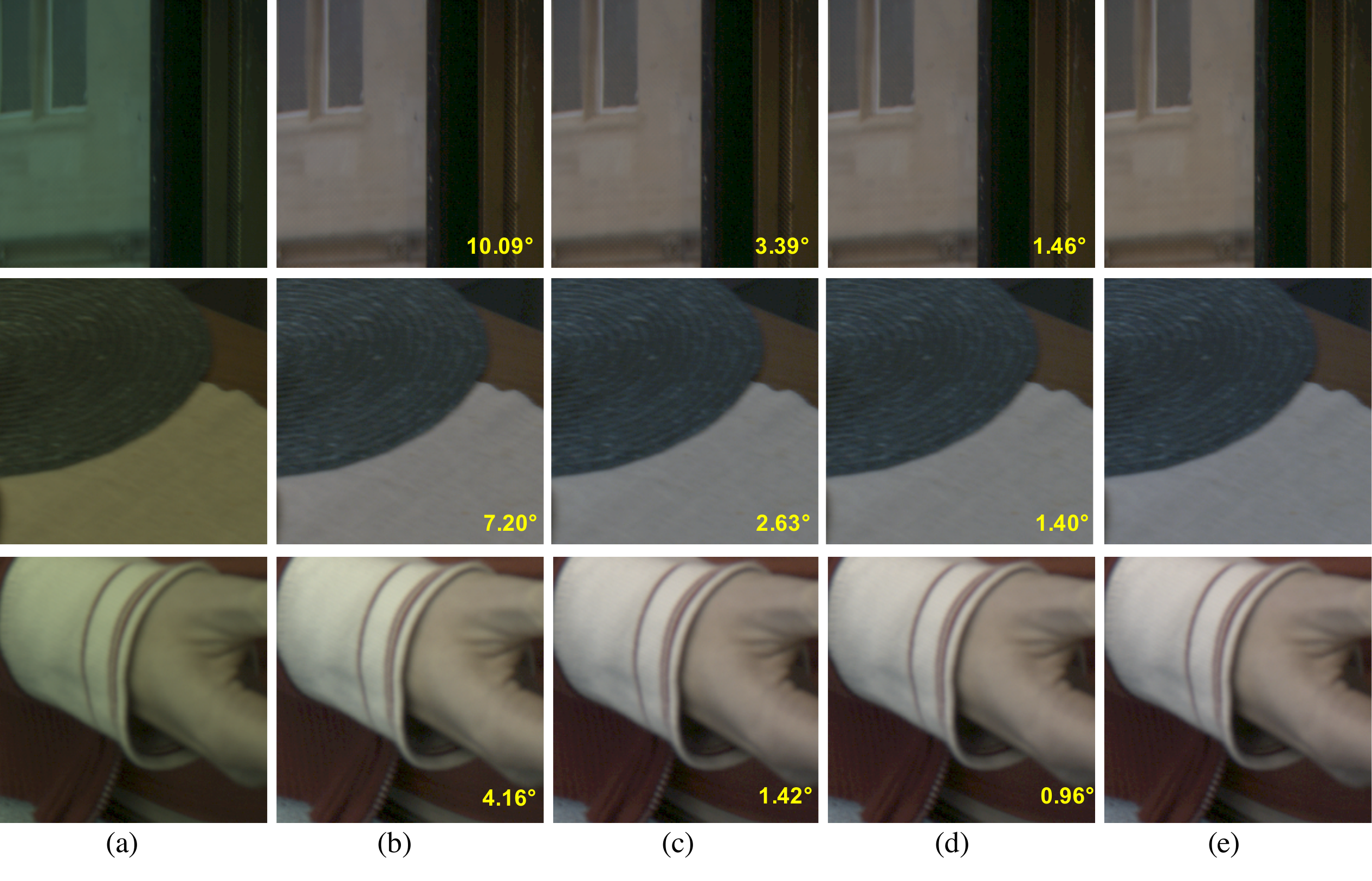}
\end{center}
   \caption{Results of different stages on two sampled patches. (a) Input patch. (b) Contextual net output. (c) Refinement output ($\mathbf{e_2}$). (d) Refinement output ($\mathbf{e_2}+\mathbf{e_3}$). (e) Ground truth. The angular error is provided at the bottom of each patch.}
\label{fig:correct_patch}
\end{figure*}

It is also interesting to observe the patches corrected early and refined stage-by-stage by the network. 
Two representative examples are visualized in Figure~\ref{fig:correct_patch}.
It is observed that the refinement net progressively improves the local estimates.

\subsubsection{Comparison of training schemes}
To further understand why the stage-wise training can achieve improvement, we explore different variants of training over the networks:
(1) training the contextual net using the VGG-16 pre-trained weights,
(2) training the {\it fc} layers of the contextual net on the basis of the off-the-shelf features obtained from the central and surround streams that are trained independently first,
(3) training the refinement net in one stage,
and (4) training the refinement net utilizing a stage-wise scheme.

The quantitative results are shown in Table~\ref{correction}.
\begin{table}[t!]
\renewcommand{\arraystretch}{1.3}
\caption{Performance comparison of different variations of the training scheme on the reprocessed Color Checker dataset}
\label{correction}
\centering
\begin{tabular}{lrrrrrr}
\hline
Method & Mean & Med & Tri & \multicolumn{1}{c}{\begin{tabular}[c]{@{}c@{}}Best\\ 25\%\end{tabular}} & \multicolumn{1}{c}{\begin{tabular}[c]{@{}c@{}}Worst\\ 25\%\end{tabular}} & \multicolumn{1}{c}{\begin{tabular}[c]{@{}c@{}}95th \\ Pct\end{tabular}} \\
\hline
Scheme (1)  & 2.31 & 1.66 & 1.82 & 0.61 &  5.23 & 6.94\\
Scheme (2) & 1.94 & 1.29 & 1.47 & 0.37 & 4.56 & 5.74 \\
Scheme (3)      & 1.83 & 1.12   & 1.29    & 0.37      & 4.39       & 5.45           \\
Scheme (4)      & \textbf{1.82} & \textbf{1.11}   & \textbf{1.29}    & \textbf{0.35}      & \textbf{4.36}       & \textbf{5.44}            \\
\hline                                                      
\end{tabular}
\end{table}
It can be seen that directly fine-tuning the contextual net initialized with the VGG-16 pre-trained model (scheme 1) decreases the performance, comparing to training the {\it fc} layers on top of the off-the-shelf {\it conv5} features (scheme 2).  
One explanation is that the network does not fully co-adapt the two streams by jointly training.
Stage-wise training benefits from treating the {\it conv} layer and the {\it fc} layers separately, and gradient diffusion can be addressed by preventing the complex co-adaptation of the feature extraction layers with the decision layer as demonstrated by~\cite{Barshan15}.
We also observe that the scheme (4) outperforms all other training schemes.
This shows that the stage-wise refinement are indeed crucial for gaining better performance.
In other words, at the early stages of training the network learns to perform the prediction by extracting coarse properties of the scene illuminant.
During following stages, finer information is gradually provided to the network and the learned intermediate features from the previous stage are re-used to perform better predictions.


\subsection{Comparison with state-of-the-arts}
Our approach is compared with previous state-of-the-art methods on the reprocessed Color Checker dataset~\cite{Gehler08} and the NUS 8-camera dataset~\cite{Cheng14}.
Most results of previous methods are directly from~\cite{Hu17,Barron17,Shi16}. 
Among them, the recent FC4~\cite{Hu17} and FFCC~\cite{Barron17} are proposed with model variants based on different backbone models or features on two datasets.
For fair comparisons, we use the results of their methods with basic components.
Tables~\ref{comarison_cc} and Table~\ref{comarison_nus8} report the metric values of the comparing methods on these two datasets, respectively.
The results of other methods are directly taken from published works.
For metric values not reported in the literature, their entries are left blank. 
\begin{table*}[t!]
\renewcommand{\arraystretch}{1.3}
\caption{Performance comparison on the reprocessed Color Checker dataset. The best three results are shown in {\color{red}{\textbf{red}}}, {\color{green}{\textbf{green}}}, and {\color{blue}{\textbf{blue}}}, respectively}
\label{comarison_cc}
\centering
\begin{tabular}{lrrrrrr}
\hline
Method                       & Mean & Med & Tri & Best-25\% & Worst-25\% & 95th Pct \\
\hline
White-Patch~\cite{Brainard86}                  & 7.55 & 5.68   & 6.35    & 1.45      & 16.12      & -               \\
Edge-based Gamut~\cite{Barnard00}             & 6.52 & 5.04   & 5.43    & 1.90      & 13.58      & -               \\
Gray-World~\cite{Buchsbaum80}                   & 6.36 & 6.28   & 6.28    & 2.33      & 10.58      & 11.3            \\
1st-order Gray-Edge~\cite{VanDeWeijer07}          & 5.33 & 4.52   & 4.73    & 1.86      & 10.03      & 11.0            \\
2nd-order Gray-Edge~\cite{VanDeWeijer07}           & 5.13 & 4.44   & 4.62    & 2.11      & 9.26       & -               \\
Shades-of-Gray~\cite{Finlayson04}               & 4.93 & 4.01   & 4.23    & 1.14      & 10.20      & 11.9            \\
Bayesian~\cite{Gehler08}                     & 4.82 & 3.46   & 3.88    & 1.26      & 10.49      & -               \\
General Gray-World~\cite{Barnard02}           & 4.66 & 3.48   & 3.81    & 1.00      & 10.09      & -               \\
Intersection-based Gamut~\cite{Barnard00}     & 4.20 & 2.39   & 2.93    & 0.51      & 10.70      & -               \\
Pixel-Based Gamut~\cite{Barnard00}            & 4.20 & 2.33   & 2.91    & 0.50      & 10.72      & 14.1            \\
Natural Image Statistics~\cite{Gijsenij11c}     & 4.19 & 3.13   & 3.45    & 1.00      & 9.22       & 11.7            \\
Bright Pixel~\cite{Joze12}                 & 3.98 & 2.61   & -       & -         & -          & -               \\
Spatio-spectral (GenPrior)~\cite{Chakrabarti12}   & 3.59 & 2.96   & 3.10    & 0.95      & 7.61       & -               \\
Cheng {\it et al.}~\cite{Cheng14}                 & 3.52 & 2.14   & 2.47    & 0.50      & 8.74       & -               \\
Corrected-Moment (19 Color)~\cite{Finlayson13}  & 3.50 & 2.60   & -       & -         & -          & 8.60            \\
Exemplar-based~\cite{Joze14}              & 3.10 & 2.30   & -       & -         & -          & -               \\
Corrected-Moment (19 Edge)~\cite{Finlayson13}   & 2.80 & 2.00   & -       & -         & -          & 6.90            \\
Regression Tree~\cite{Cheng15}              & 2.42 & 1.65   & 1.75    & 0.38      & 5.87       & -               \\
NetColorChecker~\cite{Lou15}             & 3.10 & 2.30   & -       & -         & -          & -               \\
CNN~\cite{Bianco17}                          & 2.36 & 1.98   & -       & -         & -          & -               \\
Oh \& Kim~\cite{Oh17}                    & 2.16 & 1.47   & 1.61    & 0.37      & 5.12       & -               \\
CCC (dist+ext)~\cite{Barron15}               & 1.95 & 1.22   & \color{blue}{\textbf{1.38}}    & 0.35     & \color{blue}{\textbf{4.76}}       & \color{green}{\textbf{5.85}}            \\
DS-Net (HpyNet+SelNet)~\cite{Shi16}       & \color{blue}{\textbf{1.90}} & \color{green}{\textbf{1.12}}   & \color{green}{\textbf{1.33}}    & \color{green}{\textbf{0.31}}      & 4.84       & \color{blue}{\textbf{5.99}}            \\
AlexNet-FC4~\cite{Hu17}                  & \color{red}{\textbf{1.77}} & \color{red}{\textbf{1.11}}   & \color{red}{\textbf{1.29}}    & \color{blue}{\textbf{0.34}}      & \color{red}{\textbf{4.29}}       & \color{red}{\textbf{5.44}}           \\
FFCC-thumb~\cite{Barron17}         & 2.01 &1.13  &\color{blue}{\textbf{1.38}}  &\color{red}{\textbf{0.30}}  &5.14 &-  \\
Ours                        & \color{green}{\textbf{1.82}}      & \color{red}{\textbf{1.11}}       & \color{red}{\textbf{1.29}}       & 0.35        & \color{green}{\textbf{4.36}}           & \color{red}{\textbf{5.44}}     \\
\hline       
\end{tabular}
\end{table*}

\begin{table*}[t!]
\renewcommand{\arraystretch}{1.3}
\caption{Performance comparison on the NUS 8-camera dataset. The best three results are shown in {\color{red}{\textbf{red}}}, {\color{green}{\textbf{green}}}, and {\color{blue}{\textbf{blue}}}, respectively}
\label{comarison_nus8}
\centering
\begin{tabular}{lrrrrrr}
\hline
Method                      & Mean                 & Med              & Tri              & Best-25\%            & Worst-25\%           & Geomean              \\
\hline
White-Patch~\cite{Brainard86}                  & 10.62                & 10.58                & 10.49                & 1.86                 & 19.45                & 8.43                 \\
Edge-based Gamut~\cite{Barnard00}            & 8.43                 & 7.05                 & 7.37                 & 2.41                 & 16.08                & 7.01                 \\
Pixel-based Gamut~\cite{Barnard00}           & 7.70                 & 6.71                 & 6.90                 & 2.51                 & 14.05                & 6.60                 \\
Intersection-based Gamut~\cite{Barnard00}    & 7.20                 & 5.96                 & 6.28                 & 2.20                 & 13.61                & 6.05                 \\
Gray-World~\cite{Buchsbaum80}                  & 4.14                 & 3.20                 & 3.39                 & 0.90                 & 9.00                 & 3.25                 \\
Bayesian~\cite{Gehler08}                     & 3.67                 & 2.73                 & 2.91                 & 0.82                 & 8.21                 & 2.88                 \\
Natural Image Statistics~\cite{Gijsenij11c}     & 3.71                 & 2.60                 & 2.84                 & 0.79                 & 8.47                 & 2.83                 \\
Shades-of-Gray~\cite{Finlayson04}              & 3.40                 & 2.57                 & 2.73                 & 0.77                 & 7.41                 & 2.67                 \\
Spatio-spectral (ML)~\cite{Chakrabarti12}        & 3.11                 & 2.49                 & 2.60                 & 0.82                 & 6.59                 & 2.55                 \\
General Gray-World~\cite{Barnard02}           & 3.21                 & 2.38                 & 2.53                 & 0.71                 & 7.10                 & 2.49                 \\
2nd-order Gray-Edge~\cite{VanDeWeijer07}         & 3.20                 & 2.26                 & 2.44                 & 0.75                 & 7.27                 & 2.49                 \\
Bright Pixel~\cite{Joze12}                & 3.17                 & 2.41                 & 2.55                 & 0.69                 & 7.02                 & 2.48                 \\
1st-order Gray-Edge~\cite{VanDeWeijer07}         & 3.20                 & 2.22                 & 2.43                 & 0.72                 & 7.36                 & 2.46                 \\
Spatio-spectral (GenPrior)~\cite{Chakrabarti12}  & 2.96                 & 2.33                 & 2.47                 & 0.80                 & 6.18                 & 2.43                 \\
Cheng {\it et al.}~\cite{Cheng14}                & 2.92                 & 2.04                 & 2.24                 & 0.62                 & 6.61                 & 2.23                 \\
CCC (dist+ext)~\cite{Barron15}              & 2.38                 & \color{blue}{\textbf{1.48}}                 & \color{blue}{\textbf{1.69}}                 & \color{green}{\textbf{0.45}}                 & 5.85                 & \color{blue}{\textbf{1.74}}                 \\
Oh \& Kim~\cite{Oh17}                    & 2.36                 & 2.09                 & -                    & -                    & \color{red}{\textbf{4.16}}                 & -                    \\
Regression Tree~\cite{Cheng15}             & 2.36                 & 1.59                 & 1.74                 & 0.49                 & 5.54                 & 1.78                 \\
DS-Net(HpyNet+SelNet)~\cite{Shi16}       & 2.24                 & \color{green}{\textbf{1.46}}                 & \color{green}{\textbf{1.68}}                 & 0.48                 & 6.08                 & \color{blue}{\textbf{1.74}}                 \\
AlexNet-FC4~\cite{Hu17}                & \color{green}{\textbf{2.12}}                 & 1.53                 & \color{red}{\textbf{1.67}}                 & 0.48                & \color{green}{\textbf{4.78}}                 & \color{green}{\textbf{1.66}}                 \\
FFCC-thumb~\cite{Barron17}         & \color{red}{\textbf{2.06}} &\color{red}{\textbf{1.39}}  &1.53  &\color{red}{\textbf{0.39}}  &\color{blue}{\textbf{4.80}} &-  \\
Ours                     & \color{blue}{\textbf{2.17}}      & 1.50      & \color{red}{\textbf{1.67}}         &  \color{blue}{\textbf{0.47}}         & 5.16           & \color{red}{\textbf{1.51}}       \\
\hline      
\end{tabular}
\end{table*}

\begin{figure*}[t!]
\begin{center}
   \includegraphics[width=1\linewidth]{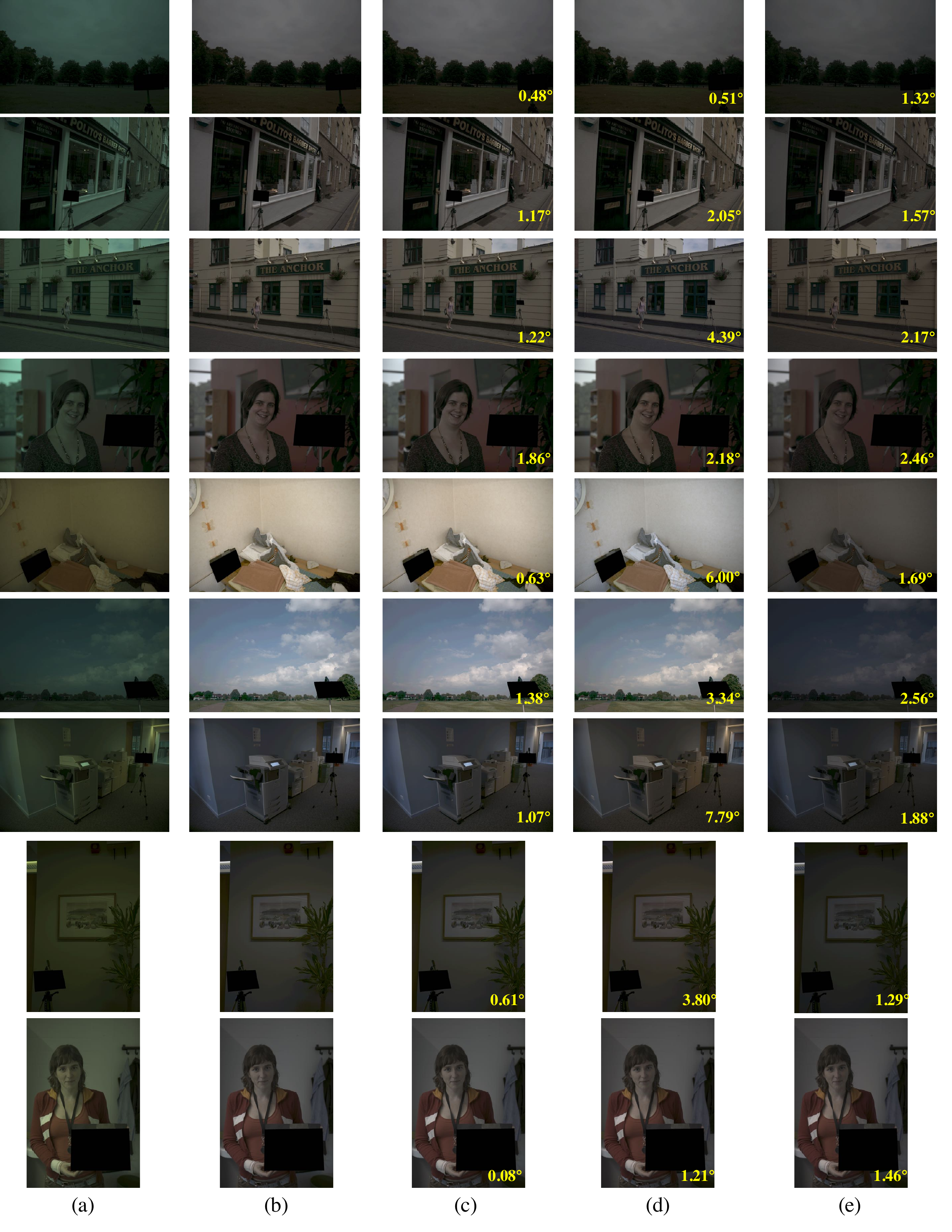}
\end{center}
   \caption{Restored images by comparing methods. (a) Input image. (b) Ground truth. (c) Ours. (d) FFCC-thumb~\cite{Barron17}. (e) AlexNet-FC4~\cite{Hu17}.}
\label{fig:comp}
\end{figure*}

As shown in the tables, we can see that the proposed network achieves competitive performance in comparison to the state-of-art methods~\cite{Barron15,Shi16,Hu17,Barron17}.
Overall our network performs better on the reprocessed Color Checker dataset than the NUS 8-camera dataset. 
The reason is that the larger size of the reprocessed Color Checker dataset facilitates our learning. 
Moreover, our approach performs similarly to AlexNet-FC4~\cite{Hu17} (training on patches and improving local estimates via confidence-weighted pooling) across some error metrics, suggesting that the use of neighbor contexts and refinement for local patches is the driving force behind our algorithm’s performance. 
However, we see a performance reduction in best-25\% error compared with FFCC-thumb~\cite{Barron17} on the two datasets, which is likely because semantic information from object detection used in FFCC-thumb favors the lowest-error of the data. 
Our experiments suggest that color constancy algorithms may benefit from much larger datasets and high-level semantic cues. 
We show restored images of our approach against two recent SOTA methods on some sample images in Figure~\ref{fig:comp}.

\section{Discussion}
\label{conclusion}
\begin{figure*}[t!]
\begin{center}
   \includegraphics[width=1\linewidth]{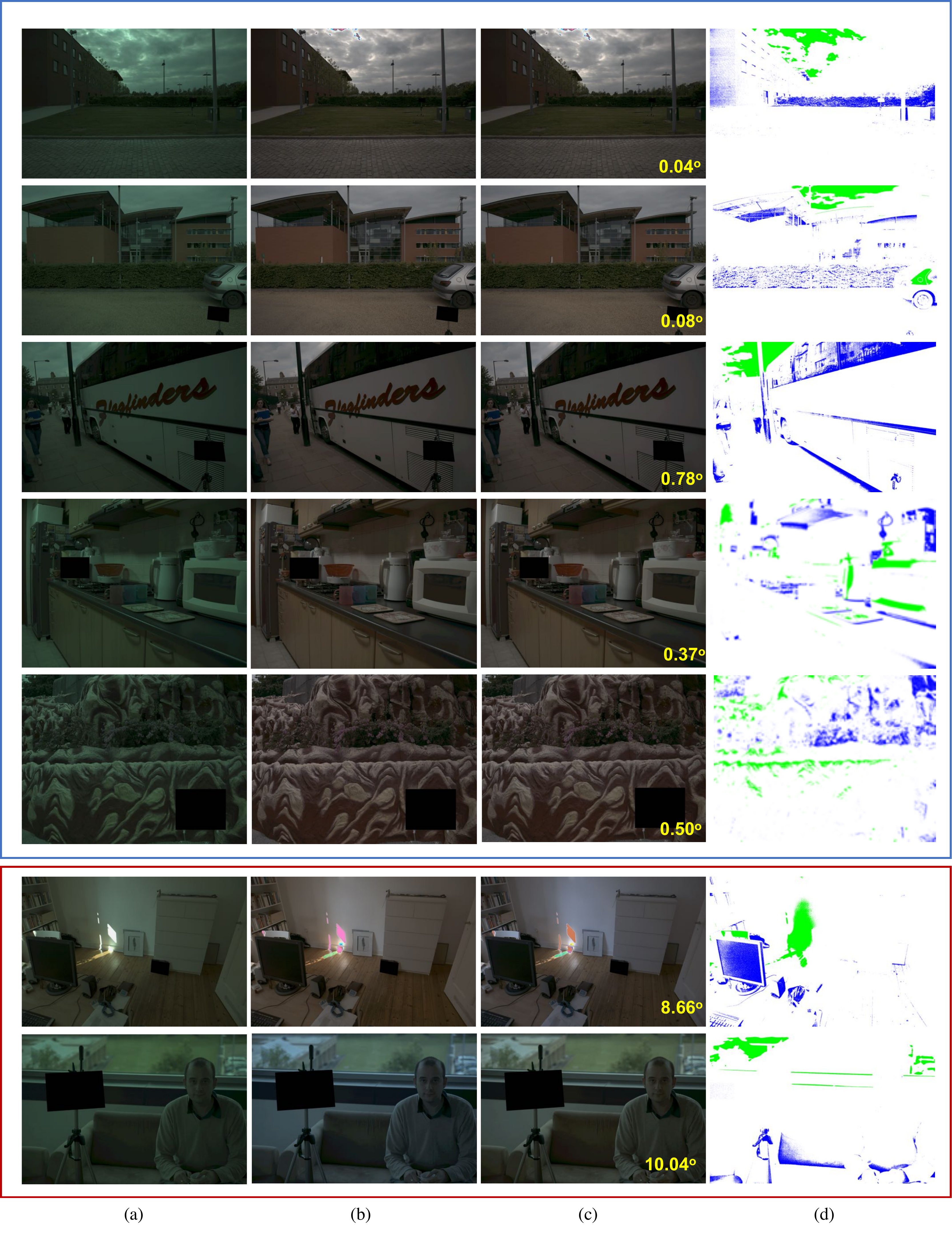}
\end{center}
   \caption{Restored images by our approach. (a) Input image. (b) Ground truth. (c) Our output. (d) Distribution of the bright and dark pixels (green: bright pixels, blue: dark pixels). Typical failure cases in the red box include glint (top) and the bright and dark pixels located far apart in the image (bottom).}
\label{fig:examples}
\end{figure*}

We propose a novel deep network for patch-based illuminant estimation.
The network is able to capture the local contextual information using the central and surround convolution units with a refinement mechanism for reevaluation of the initial estimates and features.
Our patch sampling method can significantly improve the performance of the network.
Learning features over the joint space of the input patch and the output illumination in conjunction with the use of the intermediate features are critical for training the network in a stage-wise fashion. 
There still exist difficult cases not handled perfectly by our approach, as shown in the red box of Figure~\ref{fig:examples}.
Comparing with the easy cases (shown in the blue box) restored by our approach, we found that non-solid color regions with glint and dark areas (especially achromatic ones), and bright and dark pixels located far apart generally lead the sampled patches to be biased with illuminant ambiguities, which are not meaningful enough to represent the scene illumination of the whole image. 
Therefore, removing noisy data from the training set, exploiting more effective contextual schemes (e.g., tuned suppression~\cite{Mely16}) or high-level semantic cues at a global level may alleviate such problems.

\ifCLASSOPTIONcaptionsoff
  \newpage
\fi

\bibliographystyle{IEEEtran}
\bibliography{IEEEabrv,cc}

\begin{thebibliography}{10}
\providecommand{\url}[1]{#1}
\csname url@samestyle\endcsname
\providecommand{\newblock}{\relax}
\providecommand{\bibinfo}[2]{#2}
\providecommand{\BIBentrySTDinterwordspacing}{\spaceskip=0pt\relax}
\providecommand{\BIBentryALTinterwordstretchfactor}{4}
\providecommand{\BIBentryALTinterwordspacing}{\spaceskip=\fontdimen2\font plus
\BIBentryALTinterwordstretchfactor\fontdimen3\font minus
  \fontdimen4\font\relax}
\providecommand{\BIBforeignlanguage}[2]{{%
\expandafter\ifx\csname l@#1\endcsname\relax
\typeout{** WARNING: IEEEtran.bst: No hyphenation pattern has been}%
\typeout{** loaded for the language `#1'. Using the pattern for}%
\typeout{** the default language instead.}%
\else
\language=\csname l@#1\endcsname
\fi
#2}}
\providecommand{\BIBdecl}{\relax}
\BIBdecl

\bibitem{Gijsenij11}
A.~Gijsenij, T.~Gevers, and J.~van~de Weijer, ``Computational color constancy:
  Survey and experiments,'' \emph{TIP}, vol.~20, no.~9, pp. 2475--2489, 2011.

\bibitem{Wang18}
T.~Wang, T.~Ritschel, and N.~Mitra, ``Joint material and illumination
  estimation from photo sets in the wild,'' in \emph{3DV}, 2018, pp. 22--31.

\bibitem{Barnard02}
K.~Barnard, L.~Martin, A.~Coath, and B.~Funt, ``A comparison of computational
  color constancy algorithms---part ii: Experiments with image data,''
  \emph{TIP}, vol.~11, no.~9, pp. 985--996, 2002.

\bibitem{Gijsenij07}
A.~Gijsenij and T.~Gevers, ``Color constancy by local averaging,'' in
  \emph{ICIAPW}, 2007.

\bibitem{Finlayson01}
G.~D. Finlayson and G.~Schaefer, ``Solving for colour constancy using a
  constrained dichromatic reflection model,'' \emph{IJCV}, vol.~42, no.~3, pp.
  127--144, 2001.

\bibitem{Cardei02}
V.~C. Cardei, B.~Funt, and K.~Barnar, ``Estimating the scene illumination
  chromaticity using a neural network,'' \emph{JOSA A}, vol.~19, no.~12, pp.
  2374--2386, 2002.

\bibitem{Xiong06}
W.~Xiong and B.~Funt, ``Estimating illumination chromaticity via support vector
  regression,'' \emph{JIST}, vol.~50, no.~4, pp. 341--348, 2006.

\bibitem{Gehler08}
P.~V. Gehler, C.~Rother, A.~Blake, T.~Minka, and T.~Sharp, ``Bayesian color
  constancy revisited,'' in \emph{CVPR}, 2008.

\bibitem{Joze14}
H.~R.~V. Joze and M.~S. Drew, ``Exemplar-based colour constancy and multiple
  illumination,'' \emph{TPAMI}, vol.~36, no.~5, pp. 860--873, 2014.

\bibitem{Bianco17}
S.~Bianco, C.~Cusano, and R.~Schettini, ``Single and multiple illuminant
  estimation using convolutional neural network,'' \emph{TIP}, vol.~26, no.~9,
  pp. 4347--4362, 2017.

\bibitem{Hu17}
Y.~Hu, B.~Wang, and S.~Lin, ``Fc4: Fully convolutional color constancy with
  confidence-weighted pooling,'' in \emph{CVPR}, 2017.

\bibitem{Barron17}
J.~T. Barron and Y.-T. Tsai, ``Fast fourier color constancy,'' in \emph{CVPR},
  2017, pp. 21--26.

\bibitem{Shi16}
W.~Shi, C.~C. Loy, and X.~Tang, ``Deep specialized network for illuminant
  estimation,'' in \emph{ECCV}, 2016.

\bibitem{Bianco15}
S.~Bianco, C.~Cusano, and R.~Schettini, ``Color constancy using cnns,'' in
  \emph{CVPRW}, 2015.

\bibitem{Barron15}
J.~T. Barron, ``Convolutional color constancy,'' in \emph{ICCV}, 2015.

\bibitem{Lou15}
Z.~Lou, T.~Gevers, N.~Hu, and M.~P. Lucassen, ``Color constancy by deep
  learning,'' in \emph{BMVC}, 2015, pp. 76--1.

\bibitem{Hansen07}
T.~Hansen, S.~Walter, and K.~R. Gegenfurtner, ``Effects of spatial and temporal
  context on color categories and color constancy,'' \emph{JOV}, vol.~7, no.~4,
  pp. 1--15, 2007.

\bibitem{Cheng14}
D.~Cheng, D.~K. Prasad, and M.~S. Brown, ``Illuminant estimation for color
  constancy: why spatial-domain methods work and the role of the color
  distribution,'' \emph{JOSA A}, vol.~31, no.~5, pp. 1049--1058, 2014.

\bibitem{Simonyan14}
K.~Simonyan and A.~Zisserman, ``Very deep convolutional networks for
  large-scale image recognition,'' arXiv:1409.1556, 2014.

\bibitem{vonKries70}
J.~von Kries, ``Influence of adaptation on the effects produced by luminous
  stimuli,'' in \emph{Sources of Color Science}.\hskip 1em plus 0.5em minus
  0.4em\relax Cambridge, Mass, USA: MIT Press, 1970, pp. 109--119.

\bibitem{Newel16}
A.~Newel, K.~Yang, and J.~Deng, ``Stacked hourglass networks for human pose
  estimation,'' in \emph{ECCV}, 2016, pp. 483--499.

\bibitem{Li16}
K.~Li, B.~Hariharan, and J.~Malik, ``Iterative instance segmentation,'' in
  \emph{CVPR}, 2016, pp. 3659--3667.

\bibitem{Barshan15}
E.~Barshan and P.~Fieguth, ``Stage-wise training: An improved feature learning
  strategy for deep models,'' in \emph{1st International Workshop on Feature
  Extraction: Modern Questions and Challenges at NIPS}, 2015, pp. 49--59.

\bibitem{Weijer07}
J.~van~de Weijer, T.~Gevers, and A.~Gijsenij, ``Edge-based color constancy,''
  \emph{TIP}, vol.~16, no.~9, pp. 2207--2214, 2007.

\bibitem{Finlayson11}
G.~Finlayson, S.~Hordley, and P.~Hubel, ``Color by correlation: a simple,
  unifying framework for color constancy,'' \emph{TPAMI}, vol.~23, no.~11, pp.
  1209--1221, 2011.

\bibitem{Krizhevsky12}
A.~Krizhevsky, I.~Sutskever, and G.~E. Hinton, ``Imagenet classification with
  deep convolutional neural networks,'' in \emph{NIPS}, vol.~1, 2012, pp.
  1097--1105.

\bibitem{Oh17}
S.~W. Oh and S.~J. Kim, ``Approaching the computational color constancy as a
  classification problem through deep learnin,'' \emph{PR}, vol.~61, pp.
  405--416, 2017.

\bibitem{Russakovsky15}
O.~Russakovsky, J.~Deng, H.~Su, J.~Krause, S.~Satheesh, S.~Ma, Z.~Huang,
  A.~Karpathy, A.~Khosla, M.~Bernstein, A.~C. Berg, and L.~Fei-Fei, ``Imagenet
  large scale visual recognition challenge,'' \emph{IJCV}, vol. 115, no.~3, pp.
  211--252, 2015.

\bibitem{Hurlbert07}
A.~Hurlbert, ``Colour constancy,'' \emph{Current Biology}, vol.~17, no.~21, pp.
  R906--R907, 2007.

\bibitem{Zagoruyko15}
S.~Zagoruyko and N.~Komodakis, ``Learning to compare image patches via
  convolutional neural networks,'' in \emph{CVPR}, 2015, pp. 4353--4361.

\bibitem{Ramakrishna14}
V.~Ramakrishna, D.~Munoz, M.~Hebert, J.~A. Bagnell, and Y.~Sheikh, ``Pose
  machines: Articulated pose estimation via inference machines,'' in
  \emph{ECCV}, 2014, pp. 33--47.

\bibitem{Jia14}
Y.~Jia, E.~Shelhamer, J.~Donahue, S.~Karayev, J.~Long, R.~Girshick,
  S.~Guadarrama, and T.~Darrell, ``Caffe: Convolutional architecture for fast
  feature embedding,'' arXiv preprint arXiv:1408.5093, 2014.

\bibitem{Brainard86}
D.~H. Brainard and B.~A. Wandell, ``Analysis of the retinex theory of color
  vision,'' \emph{JOSA A}, vol.~3, no.~10, pp. 1651--1661, 1986.

\bibitem{Barnard00}
K.~Barnard, ``Improvements to gamut mapping colour constancy algorithms,'' in
  \emph{ECCV}, 2000, pp. 390--403.

\bibitem{Buchsbaum80}
G.~Buchsbaum, ``A spatial processor model for object colour perception,''
  \emph{Journal of the Franklin institute}, vol. 310, no.~1, pp. 1--26, 1980.

\bibitem{VanDeWeijer07}
J.~V.~D. Weijer, T.~Gevers, and A.~Gijsenij, ``Edge-based color constancy,''
  \emph{TIP}, vol.~16, no.~9, pp. 2207--2214, 2007.

\bibitem{Finlayson04}
G.~D. Finlayson and E.~Trezzi, ``Shades of gray and colour constancy,'' in
  \emph{CIC}, 2004, pp. 37--41.

\bibitem{Gijsenij11c}
A.~Gijsenij and T.~Gevers, ``Color constancy using natural image statistics and
  scene semantics,'' \emph{TPAMI}, vol.~33, no.~4, pp. 687--698, 2011.

\bibitem{Joze12}
H.~R.~V. Joze, M.~S. Drew, G.~D. Finlayson, and P.~A.~T. Rey, ``The role of
  bright pixels in illumination estimation,'' in \emph{CIC}, vol.~1, 2012, pp.
  41--46.

\bibitem{Chakrabarti12}
A.~Chakrabarti, K.~Hirakawa, and T.~Zickler, ``Color constancy with
  spatio-spectral statistics,'' \emph{TPAMI}, vol.~34, no.~8, pp. 1509--1519,
  2012.

\bibitem{Finlayson13}
G.~D. Finlayson, ``Corrected-moment illuminant estimation,'' in \emph{ICCV},
  2013, pp. 1904--1911.

\bibitem{Cheng15}
D.~Cheng, B.~Price, S.~Cohen, and M.~S. Brown, ``Effective learning-based
  illuminant estimation using simple features,'' in \emph{CVPR}, 2015, pp.
  1000--1008.

\bibitem{Mely16}
D.~A. Mely and T.~Serre, ``Opponent surrounds explain diversity of contextual
  phenomena across visual modalities,'' bioRxiv, p.070821, 2016.

\end{thebibliography}




\end{document}